\documentclass[runningheads]{llncs}

\usepackage{eccv}

\usepackage{eccvabbrv}

\usepackage{graphicx}
\usepackage{booktabs}
\usepackage{wrapfig}
\usepackage[table]{xcolor}
\usepackage{caption}

\usepackage[accsupp]{axessibility}  

\usepackage{hyperref}

\usepackage{orcidlink}

\begin{document}

\title{Coarse-to-fine Framework for Generative MEF via Implicit Neural Representation} 

\titlerunning{Coarse-to-fine Framework for Generative MEF via INR}

\author{Sangmin Han\inst{1,2}\orcidlink{0009-0001-6267-7290} \and
Jinho Kim\inst{1}\orcidlink{0009-0000-2055-8683} \and
Jinwoo Kim\inst{1}\orcidlink{0009-0001-3250-1788} \and \\
Dongyoung Kim\inst{1}\orcidlink{0009-0000-6414-2380} \and
Seon Joo Kim\inst{1}\orcidlink{0000-0001-8512-216X}
}

\authorrunning{S.Han et al.}

\institute{
    $^1$Yonsei University \,
    $^2$AI Lab, CTO Division, LG Electronics
}

\maketitle

\begin{center}
    \vspace{-3mm}
    \footnotesize \textbf{Project Page:} \url{https://sangmin213.github.io/LIIFusion/}
\end{center}

\begin{center}
    \centering
    \includegraphics[width=\linewidth]{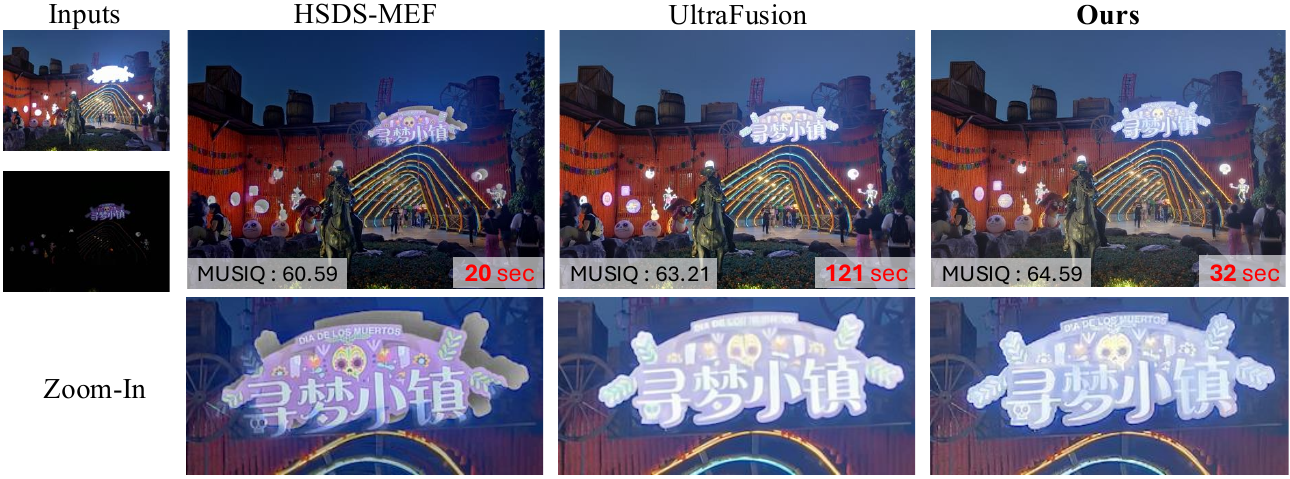}
    \captionof{figure}{
        Our coarse-to-fine generative MEF framework preserves the natural synthesis of generative MEF (color tone and reduced ghosting) while achieving detailed structural preservation comparable to conventional MEF. It is about $3.5\times$ faster than the previous generative MEF method (Ultrafusion~\cite{chen2025ultrafusion}) and runs close to conventional MEF (HSDS-MEF~\cite{wu2024hybridsupervised}); speed is measured on a single NVIDIA RTX A5000 with a $1228\times1638$-resolution image. The first-row insets are center-cropped.
        }
    \label{fig:teaser}
    \vspace{-1pt}
\end{center} 

\begin{abstract}
Multi-exposure fusion (MEF) expands the luminance range beyond what a single exposure can capture.
Combining images taken at different exposure levels requires handling geometric differences while naturally merging their complementary brightness information.
It often demands generative completion where details are missing.
Diffusion-based generative methods address these challenges, however, they are computationally expensive and struggle to preserve fine structures in saturated regions.
We propose LIIFusion, a coarse-to-fine framework that balances fusion quality and efficiency in generative MEF.
The coarse stage performs low resolution generative fusion, enhanced by an adaptive exposure correction that recovers structure lost in saturated over-exposed areas.
The fine stage adapts a local implicit image function into a multi-exposure fusion function: conditioned on the HR OE/UE sources and the coarse output, it queries arbitrary target coordinates and fuses source evidence regard- less of the HR input resolution.
LIIFusion achieves up to 3.5$\times$ speed-up over existing generative methods while maintaining or improving structural fidelity and perceptual quality.
We believe this framework provides an effective pathway toward making generative MEF more practical in real-world applications.
\end{abstract}    
\section{Introduction}
\label{sec:intro}
A fundamental challenge in computational photography is capturing scenes with a high dynamic range (HDR). Due to the limited dynamic range of standard sensors, a single exposure is often insufficient to capture the full range of scene illumination, resulting in over-exposed backgrounds or under-exposed foregrounds. A common solution to this problem is multi-exposure fusion (MEF). This technique involves capturing a sequence of images at varying exposure levels and synthesizing them into a single high-quality image that preserves details across all regions.

The core objective of MEF is to blend salient, well-exposed regions from the input sequence while maintaining natural tonal transitions. Traditional approaches~\cite{mertens2007exposure} tackled this using hand-crafted, multi-scale pyramid fusion, which relies on pixel-level measures such as contrast, saturation, and well-exposedness. However, these low-level heuristics often produce halo artifacts and, crucially, lack the semantic awareness to distinguish important foreground details from background texture. To address these limitations, modern deep learning methods~\cite{xu2020mef, liang2022fusion, jiang2023meflut, wu2024hybridsupervised} have improved robustness by learning end-to-end fusion strategies. These models often operate on rich, feature-level representations and may adopt multi-scale designs for computational efficiency.

\begin{wraptable}[10]{r}{0.5\linewidth}
  \vspace{-2\baselineskip} 
  \centering
  \caption{Comparison between conventional MEF and generative MEF.}
  \label{tab:mef_comparison}
  \renewcommand{\arraystretch}{1.3}
  \resizebox{\linewidth}{!}{%
    \begin{tabular}{l|c|cc}
      \toprule
      & \textbf{Conventional} & \multicolumn{2}{c}{\textbf{Generative MEF}} \\
      & \textbf{MEF} & \textbf{Patch-wise} & \textbf{Coarse-to-fine} \\
      \midrule \midrule
      Models & CNN, Transformer & Diffusion & Diffusion, LIIF \\
      \hline
      Dynamic Range & Limited (3-4 stops) & \multicolumn{2}{c}{Extended (9 stops)} \\
      \hline
      Motion Handling & Static / Mild & \multicolumn{2}{c}{Dynamic} \\
      \hline
      Fusion Behavior & Regressive & Probabilistic & Both \\
      \hline
      Speed & Fast ($\sim$minutes) & Slow ($\sim$hours) & Fast ($\sim$minutes) \\
      \bottomrule
    \end{tabular}%
  }
  \vspace{\baselineskip} 
\end{wraptable}

However, both conventional and early learning-based approaches face significant challenges in real-world scenarios. They frequently exhibit ghosting artifacts when subject motion occurs between exposures and fail in scenes with extreme dynamic ranges, where large regions become saturated and require generative synthesis rather than simple blending. Recent works~\cite{chen2025ultrafusion, zhu2025flexible} have introduced diffusion-based generative priors that reinterpret MEF as a guided inpainting task. While these models effectively suppress ghosting and synthesize realistic structures in severely over- or under-exposed regions, they introduce a new set of trade-offs as summarized in \cref{tab:mef_comparison}. Specifically, high-resolution fusion is computationally expensive because diffusion models operate at fixed resolutions (e.g., $512\times512$) and require patch-wise sampling, and purely generative fusion can still yield structural uncertainty or unstable tones when guidance is weak in saturated areas.


To overcome these limitations, we propose \textbf{LIIFusion}, a coarse-to-fine generative framework that harmonizes a diffusion prior at low-resolution with a high-resolution implicit refinement module. In the coarse stage, LIIFusion performs generative fusion on downsampled exposure pairs, operating near the diffusion model’s native pre-training scale to maximize efficiency and fully exploit its inpainting priors. This design lets the diffusion model produce a single low-resolution fused image with a minimal number of forward passes, avoiding expensive patch-wise sampling. Within this stage, we introduce an adaptive exposure correction mechanism, which conditions the input to yield structurally reliable coarse estimates even in saturated regions. 
In the fine stage, we employ an implicit neural representation (INR) as a multi-exposure conditional fusion function rather than as a single-image super-resolution decoder.
This module extracts LIIF-style features~\cite{chen2021liif} from the low-resolution fusion and uses lightweight encoders for the high-resolution exposures, enabling efficient multi-resolution fusion. 
Because the function is queried in continuous coordinates and conditioned directly on the HR OE/UE sources, the same learned fusion function can integrate exposure evidence from HR inputs of arbitrary resolution.
By integrating these complementary signals at the coordinate level, the INR reconstructs a high-quality, continuous fused image. To the best of our knowledge, this constitutes the first application of INR to the task of multi-exposure fusion.

We conduct extensive quantitative and qualitative evaluations on both static and dynamic multi-exposure benchmarks. Results demonstrate that LIIFusion achieves a 3.5$\times$ speed-up over existing generative MEF methods while maintaining superior quantitative performance and perceptual quality. We believe this framework bridges the critical gap between generative fidelity and high-resolution efficiency, paving the way for the practical deployment of generative MEF in real-world imaging pipelines.

\section{Related works}
\label{sec:related_work}

\textbf{High dynamic range imaging.} 
HDR imaging aims to capture the full luminance range of real-world scenes, overcoming the sensor limitations of standard cameras such as pixel saturation and narrow dynamic range. Existing approaches generally fall into two categories: HDR reconstruction, which restores physical radiance, and Multi-Exposure Fusion (MEF), which targets perceptual quality. HDR reconstruction methods~\cite{kong2024safnet,tel2023alignment,liu2022ghost,chen2022attention,kalantari2017deep,prabhakar2020hdrcnn,song2022transhdr,wu2018deephdr,xu2024hdrflow,zheng2013hybrid,yan2020nonlocal,yan2019attention,zhu2024zeroshot,li2025afunet,wang2025lediff,bemana2025bracket} recover absolute scene radiance maps from Low Dynamic Range (LDR) inputs. While precise, these methods typically require camera response function (CRF) inversion and subsequent tone-mapping for display, which can introduce color distortion or dynamic range compression artifacts. 
\newline
\\
\textbf{Multi-exposure fusion (MEF).} 
In contrast, MEF~\cite{mertens2007exposure,ma2019deep,ma2017robust} directly fuses the LDR sequence into a perceptually high-quality image, bypassing explicit radiance recovery. Traditional MEF methods~\cite{li2017detail,li2020fast,zheng2015superpixel,prabhakar2019deghosting} relied on hand-crafted weights or multi-scale blending, often struggling with ghosting artifacts in dynamic scenes. Deep learning-based approaches~\cite{ram2017deepfuse,xu2020mef,xu2020u2fusion,jiang2023meflut,wu2024hybridsupervised,zhao2024vlfusion} have significantly improved robustness to misalignment and detail preservation. Most recently, diffusion-based methods~\cite{chen2025ultrafusion,zhu2025flexible} have leveraged generative priors to hallucinate plausible structures in saturated regions, treating MEF as a guided inpainting task. However, to handle high-resolution inputs, these diffusion models rely on patch-wise inference. This not only incurs prohibitive computational costs due to repeated sampling but can also lead to spatial inconsistencies at patch boundaries or unstable structure when guidance is weak in saturated regions. In this work, we address these bottlenecks by decoupling the generative prior from the output resolution and pairing it with deterministic refinement.
\newline
\\
\textbf{Implicit neural representations (INR).} 
INR~\cite{barron2021mipnerf,cao2023ciaosr,fang2024cycleinr,feng2024pienerf,ke2024improved,kim2024arbitrary,lee2022lte,peng2021animatable,pumarola2021dnerf,yuan2022sobolev,sabour2023robustnerf} models visual signals as continuous functions mapping spatial coordinates to signal values. Originally popularized for 3D view synthesis in NeRF~\cite{mildenhall2020nerf}, INRs have been successfully adapted to 2D tasks such as super resolution (LIIF~\cite{chen2021liif}), where they reconstruct fine details from discrete features via local implicit functions. Recently, this paradigm has extended to other low-level vision tasks. For instance, CoLIE~\cite{chobola2024colie} utilizes INRs for context-aware low-light enhancement, while INR-st~\cite{kim2022controllable} enables controllable style transfer via test-time training. Despite these advances, INR has not yet been explored in the context of multi-exposure fusion. We hypothesize that the continuous nature of INRs is uniquely suited for MEF, as it allows for the seamless integration of global generative priors (from the coarse stage) with high-frequency texture details (from the fine stage) at arbitrary coordinates.

\begin{figure*}[t]
    \centering
    \includegraphics[width=\linewidth]{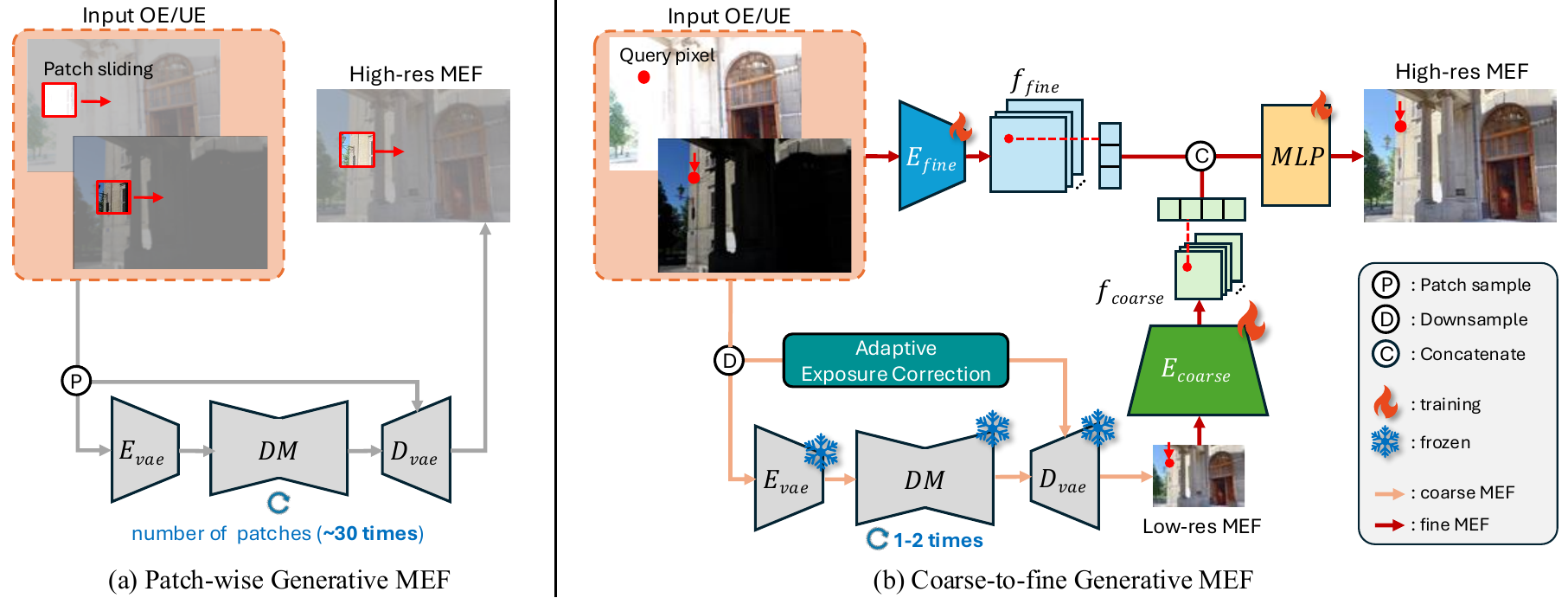}
    \caption{\textbf{Overview of LIIFusion pipeline.} (a) Previous generative MEF models operate in a patch-wise manner, requiring many sampling passes. (b) Our proposed coarse-to-fine framework, LIIFusion, performs low-resolution generative fusion with adaptive exposure correction to obtain a structurally reliable result, minimizing diffusion usage to up to 2 passes depending on the image resolution ratio, followed by a LIIF-based fine stage that conducts resolution-agnostic, pixel-wise detail fusion.} 
    \label{fig:LIIFusion_overview}
\end{figure*}
\section{Methodology}
In this section, we describe our proposed coarse-to-fine framework for generative multi-exposure fusion (MEF). We first define the notations used throughout this paper. Let $I_{oe}\in\mathbb{R}^{H\times W\times 3}$ and $I_{ue}\in\mathbb{R}^{H\times W\times 3}$ denote the high-resolution (HR) over-exposed (OE) and under-exposed (UE) input images, respectively. Similarly, let $I_{oe}^{LR}\in\mathbb{R}^{h\times w\times 3}$ and $I_{ue}^{LR}\in\mathbb{R}^{h\times w\times 3}$ denote their low-resolution (LR) counterparts obtained via bicubic downsampling, where $H > h$ and $W > w$. 

The generative MEF model $g$ operates on these low-resolution inputs to produce a coarse fused image $I_\text{mef}^{LR}$. Subsequently, our fine stage implicit neural representation (INR) model employs two encoders: a coarse feature encoder $E_\text{coarse}$ processing $I_\text{mef}^{LR}$, and a fine feature encoder $E_\text{fine}$ processing the HR exposure inputs $\{I_{oe}, I_{ue}\}$. The final output of our framework is the high-resolution fused image $I_\text{mef}^{HR}\in\mathbb{R}^{H\times W\times 3}$.

\subsection{Preliminary}
\label{subsec:preliminary}
\subsubsection{UltraFusion}
UltraFusion~\cite{chen2025ultrafusion}, which serves as the backbone for our coarse stage, is a diffusion-based generative MEF framework. Unlike conventional deep learning approaches that directly learn pixel-level fusion mappings, UltraFusion formulates MEF as a guided inpainting problem in the latent space. It leverages the generative prior of a pre-trained diffusion model to synthesize structurally consistent and artifact-free results.

Given an exposure pair $\{I_{oe}, I_{ue}\}$, UltraFusion first estimates the optical flow $F_{oe \rightarrow ue}$ between the inputs using RAFT~\cite{teed2020raft} to obtain dense motion fields. The UE image is then warped according to the flow to produce an aligned version $\tilde{I}_{ue}=W(I_{ue}, F_{oe\rightarrow ue})$ that spatially matches the OE image where $W(\cdot)$ is a backward warping operation. This alignment mitigates ghosting and establishes reliable correspondence for fusion. As illustrated in \cref{fig:LIIFusion_overview}(a), the aligned exposures $\{I_{oe}, \tilde{I}_{ue}\}$ are typically divided into patches matching the model’s pre-training resolution (e.g., $512\times512$) and processed independently. Within each patch, the OE image acts as the reference, while information from $\tilde{I}_{ue}$ guides the reconstruction of saturated regions. Finally, a VAE decoder reconstructs the fused RGB image from the latent representations with patch guidance. Through this design, UltraFusion combines the semantic reconstruction power of diffusion models with structural guidance, providing a strong foundation for generative fusion.

\subsubsection{Local implicit image function (LIIF)}
The Local implicit image function (LIIF)~\cite{chen2021liif} represents an image as a continuous function mapping spatial coordinates to RGB values. Given an input image $I$, an encoder $E$ extracts a latent feature map $M = E(I) \in \mathbb{R}^{H \times W \times D}$, where $H$ and $W$ denote the spatial dimensions and $D$ is the feature dimension. The image coordinates are normalized within $[-1, 1]$ for both axes so that any spatial location can be represented in a scale-independent manner. A local feature vector $z \in \mathbb{R}^D$ is obtained by bilinear interpolation near a query coordinate $x = [x_h, x_w] \in [-1, 1]^2$ over the feature map $M$.

A lightweight MLP decoder $f_\theta$, parameterized by $\theta$, predicts the RGB value $s \in \mathbb{R}^3$ at the coordinate $x$ as:
\begin{equation}
    s = f_\theta(z, [x, c]),
    \label{eq:liif}
\end{equation}
where $c = [c_h, c_w]$ represents the relative size of the query pixel, referred to as the \textit{cell}. The cell encodes the spatial extent of each query and allows the function to adapt to varying output resolutions.

This implicit representation enables the model to process and integrate image information across different resolutions in a continuous manner. In our framework, we leverage this property to jointly utilize the coarse generative fusion result $I_\text{mef}^{LR}$ and the high-resolution exposure images $\{I_{oe}, I_{ue}\}$, allowing the model to predict the RGB value of each coordinate in the target high-resolution fused image $I_\text{mef}^{HR}$ through unified multi-resolution image fusion.

\subsection{LIIFusion}
\label{subsec:framework}
The proposed model, \textbf{LIIFusion}, performs generative MEF in a coarse-to-fine manner. The framework consists of two complementary stages: a coarse fusion stage based on UltraFusion~\cite{chen2025ultrafusion} for exposure harmonization and ghost removal, and a fine fusion stage that reconstructs high-resolution details using an implicit representation network.
\newline
\\
\textbf{Coarse fusion.}
In the coarse stage, LIIFusion adopts UltraFusion as the generative prior to perform low-resolution fusion efficiently. Given the high-resolution exposure pair $\{I_{oe}, I_{ue}\}$, both images are first downsampled to obtain $\{I_{oe}^{LR}, I_{ue}^{LR}\}$. Optical flow is estimated using RAFT~\cite{teed2020raft}, yielding dense motion field $F_{oe\rightarrow ue}$. The field and its bicubic-downsampled counterpart are used to warp the high- and low-resolution under-exposed images, respectively, producing $\{\tilde{I}_{ue}, \tilde{I}_{ue}^{LR}\}$. The aligned low-resolution pair $\{I_{oe}^{LR}, \tilde{I}_{ue}^{LR}\}$ is then processed by the diffusion model $g$ to produce a coarse fused image:
\begin{equation}
    I_\text{mef}^{LR} = g(I_{oe}^{LR}, \tilde{I}_{ue}^{LR}).
\end{equation}
\newline
\\
\textbf{Fine fusion.}
The fine stage reconstructs the target high-resolution fused image $I_\text{mef}^{HR}$ by integrating features across resolutions. The coarse feature encoder $E_\text{coarse}$ extracts representations $f_\text{coarse} = E_\text{coarse}(I_\text{mef}^{LR})$ from the generative output, while the fine feature encoder $E_\text{fine}$ encodes the concatenated high-resolution exposures as $f_\text{fine} = E_\text{fine}(I_{oe}, \tilde{I}_{ue})$.

To predict the RGB value at an arbitrary query coordinate $x$ in the high-resolution domain, we first obtain local feature vectors $z_{coarse}$ and $z_{fine}$ by performing bilinear interpolation on $f_\text{coarse}$ and $f_\text{fine}$ at location $x$. These vectors are concatenated and decoded by a lightweight MLP $f_\theta$:
\begin{equation}
    s = f_\theta([z_\text{coarse}, z_\text{fine}], [x, c]).
\end{equation}
The final image $I_\text{mef}^{HR}$ is formed by querying all coordinates in the target resolution. The network is trained using an $L_1$ loss against the ground truth $I_{GT}\in\mathbb{R}^{H\times W\times 3}$:
\begin{equation}
    \mathcal{L}_\text{fusion} = \| I_\text{mef}^{HR} - I_{GT} \|_1.
\end{equation}
This design allows the diffusion-based coarse stage to handle exposure inconsistencies and motion artifacts globally, while the implicit representation refines geometric and texture details locally at full resolution.

\subsection{Adaptive exposure correction}
\label{subsec:exposure_control}

Diffusion-based generative fusion models typically perform exposure blending via guided inpainting in the latent space~\cite{zhu2025flexible}. UltraFusion~\cite{chen2025ultrafusion} employs a fidelity guidance stage within the VAE decoder, utilizing input exposures to ensure structural consistency. While effective, this design struggles when the over-exposed (OE) input is severely saturated, rendering the guidance unreliable. This issue is exacerbated at low resolutions, where the model is required to distinguish fine details within a limited receptive field. To address this, we propose an adaptive exposure correction module that refines the OE image prior to its use in fidelity guidance, thereby providing a stable structural cue for the generative decoder.
\newline
\\
\textbf{Exposure difference estimation.}
Given exposure pairs $I_{oe}^{LR}, I_{ue}^{LR}$, we first convert them to the LAB color space and extract their luminance channels $L_{oe}^{LR}, L_{ue}^{LR} \in [0, 1]^{h\times w\times3}$. An exposure difference map is computed as:
\begin{equation}
    L_\text{diff}^{LR} = \text{min}(\text{max}(L_{oe}^{LR} - L_{ue}^{LR}, 0), 1),
\end{equation}
where $L_\text{diff}^{LR}$ highlights saturated regions exhibiting large luminance disparities. We explicitly clip values to $[0,1]$ for numerical stability.
\newline
\\
\textbf{Adaptive brightness adjustment.}
A per-pixel weight map $\textbf{W}$ is generated to modulate excessively bright regions in $I_{oe}^{LR}$:
\begin{equation}
    \textbf{W} = (1 - \alpha \cdot L_\text{diff}^{LR})^{1/2.2},
    \label{eq:aec_weight_map}
\end{equation}
where $\alpha$ is a scalar controlling the strength of compensation and the exponent $1/2.2$ applies perceptual gamma correction to smooth the adjustment curve~\cite{bemana2025bracket}.
The adjusted OE image is then obtained via:
\begin{equation}
    I_{oe}^{\prime{LR}} = I_{oe}^{LR} \odot \textbf{W},
\end{equation}
where $\odot$ denotes element-wise multiplication. Finally, $I_{oe}^{\prime{LR}}$ replaces the original $I_{oe}^{LR}$ during the fidelity guidance stage of the generative fusion pipeline.
\section{Experiments}
\label{sec: experiments}

\subsection{Experimental setting}
\textbf{Datasets.}
We construct our training dataset from the SICE dataset~\cite{cai2018learning} by randomly selecting under- and over-exposed image pairs from each exposure bracket. For evaluation, we assess the performance of our model on both static and dynamic datasets, including MEFB~\cite{Zhang2021} with 100 static exposure pairs, the UltraFusion Benchmark~\cite{chen2025ultrafusion} containing 100 real-captured under/over-exposed pairs with larger exposure differences, and the RealHDRV dataset~\cite{Shu2024} that captures 50 dynamic scenes with diverse motions.
\newline
\\
\noindent \textbf{Implementation details.}
Since the SICE dataset contains static scenes, we follow UltraFusion’s synthetic setup by multiplying an occlusion mask extracted from Vimeo-90K~\cite{xue2019video} with the UE images to simulate dynamic misalignment. SICE label data is used as ground truth, and we downsample them to random scale to obtain corresponding triplet $\{I_{oe}, I_{ue}, I_\text{mef}^{LR}\}$ for training.  
The fine encoder $E_\text{fine}$ receives $19\times19$ local patches around each target pixel from $I_{oe}$ and $I_{ue}$ to encode local exposure features, implemented as a 4-layer CNN. We adopt a SwinIR~\cite{liang2021swinir} backbone for the coarse encoder $E_\text{coarse}$ such as convention of LIIF~\cite{chen2021liif}.  
We use the Adam optimizer with a learning rate of $1\times10^{-4}$ and a batch size of 16. The model is trained for 500 epochs on a single NVIDIA H100 GPU, requiring approximately 22 hours for full convergence.
\newline
\\
\noindent \textbf{Evaluation metrics.}
To evaluate both structural preservation and perceptual naturalness, we employ MEF-SSIM~\cite{ma2015perceptual} as a structure aware fusion metric, along with four non reference image quality assessment (NRIQA) metrics: MUSIQ~\cite{ke2021musiq}, PAQ2PIQ~\cite{ying2020patches}, DeQA-Score~\cite{you2025teaching}, and HyperIQA~\cite{su2020blindly}, which measure perceptual quality.  
MEF aims to produce a content-complete, well tone-mapped image, yet there is no unique ground truth for what constitutes a “good” tone mapping; therefore, we use NRIQA to assess perceptual naturalness while relying on MEF-SSIM to verify structural preservation.  
Using these metrics, we quantitatively compare LIIFusion with existing conventional and generative MEF models across the aforementioned datasets.  
We also report total inference time per dataset, as well as the number of parameters and FLOPs for a single image, to assess the computational efficiency of our approach compared with patch-wise generative MEF methods.

\begin{table*}[t]
    \centering
    \caption{Quantitative comparison on the dynamic RealHDRV dataset~\cite{Shu2024} and UltraFusion Benchmark~\cite{chen2025ultrafusion}. \textbf{Bold} denote the best performance and \underline{underlined} indicate the second best. Total inference time for each dataset is measured on a single NVIDIA RTX A5000.}
    \resizebox{\linewidth}{!}{%
        \begin{tabular}{l|ccccc|ccccc}
        \toprule
                               & \multicolumn{5}{c|}{RealHDRV (50 scenes)}                                                                                                  
                               & \multicolumn{5}{c}{UltraFusion Benchmark (100 scenes)}                                                                                       \\
            \multicolumn{1}{c|}{Model}  & MUSIQ$\uparrow$ & DeQA-Score$\uparrow$ 
                                        & PAQ2PIQ$\uparrow$ & HyperIQA$\uparrow$ & Time$\downarrow$ 
                                        & MUSIQ$\uparrow$ & DeQA-Score$\uparrow$ 
                                        & PAQ2PIQ$\uparrow$ & HyperIQA$\uparrow$ & Time$\downarrow$ \\ 
                               \midrule \midrule
        Defusion~\cite{liang2022fusion}               & 56.38 & 3.2867 & 68.31 & 0.4838 & 2 min    
                               & 60.11 & 3.3529 & 71.83 & 0.5440 & 6 min    \\
        MEF-LUT~\cite{jiang2023meflut}                 & 62.42 & 3.2864 & 70.04 & 0.5020 & 4 sec   
                               & 64.06 & 3.2859 & 71.80 & 0.5103 & 8 sec  \\
        HSDS-MEF~\cite{wu2024hybridsupervised}               & 61.82 & 3.6045 & 71.14 & 0.5055 & 18 min    
                               & 65.23 & 3.6662 & 73.77 & 0.5786 & 46 min    \\
        UltraFusion~\cite{chen2025ultrafusion}            & \underline{67.54} & \textbf{3.8998} 
                               &\underline{73.39} & \underline{0.5834} & 101 min    
                               & \underline{68.40} & \textbf{4.0123} & \underline{75.18} & \underline{0.6214} & 203 min    \\
        Ours                   & \textbf{69.52} & \underline{3.8908}
                               & \textbf{74.06} & \textbf{0.6175} & 27 min
                               & \textbf{70.19} & \underline{3.9807}    
                               & \textbf{75.59} & \textbf{0.6467}  & 59 min \\
                               \bottomrule
        \end{tabular}%
    }
    \label{tab:realhdrv_ultrabench}
    \vspace{-1pt}
\end{table*}
\begin{table*}[t]
    \centering
    \begin{minipage}[t]{0.57\linewidth}
        \centering
        \captionof{table}{Quantitative comparison on the static MEFB dataset~\cite{Zhang2021}.}
        \resizebox{\linewidth}{!}{%
            \begin{tabular}{l|cccc}
            \toprule
                                        & \multicolumn{4}{c}{MEFB (100 scenes)} \\
            \multicolumn{1}{c|}{Model}  & MUSIQ$\uparrow$ & PAQ2PIQ$\uparrow$ 
                                        & HyperIQA$\uparrow$ & MEF-SSIM$\uparrow$ \\ 
                                   \midrule \midrule
            MEF-GAN~\cite{xu2020mef}                & 51.28 & 70.58 & 0.3974   & 0.7197   \\
            U2Fusion~\cite{xu2020u2fusion}               & 64.09 & 72.34 & 0.5436   & 0.9212   \\
            Defusion~\cite{liang2022fusion}               & 62.70 & 70.82 & 0.5454   & 0.9102   \\
            MEF-LUT~\cite{jiang2023meflut}                 & 65.71 & 71.21 & 0.5267   & 0.9005   \\
            HSDS-MEF~\cite{wu2024hybridsupervised}               & 66.81 & 72.60 & 0.6026   & \textbf{0.9367}   \\
            UltraFusion~\cite{chen2025ultrafusion}            & 68.14 & 73.45 & 0.6310   & 0.9266   \\
            Ours                   & \textbf{68.78} & \textbf{73.69} & \textbf{0.6418}  & 0.9201   \\ 
            \bottomrule
            \end{tabular}%
        }
        \vspace{-1pt}
        \label{tab:mefb}
    \end{minipage}
    \hfill
    \begin{minipage}[t]{0.40\linewidth}
        \centering
        \captionsetup{font=footnotesize}
        \captionof{table}{FLOPs and parameter comparison. FLOPs are measured on a single $1988\times1326$ image from UltraFusion Benchmark.}
        \vspace{-8pt}
        \resizebox{0.7\linewidth}{!}{%
            \begin{tabular}{l|cc}
            \toprule
            \multicolumn{1}{c|}{Model}  & Parameters & TFLOPs \\ 
                                   \midrule \midrule
            MEF-GAN               & 0.488 M       & 2.483  \\
            U2Fusion             & 0.659 M       & 3.477   \\
            Defusion             & 7.874 M     & 0.640   \\
            MEF-LUT               & 0.061 M        & 0.003     \\
            HSDS-MEF             & 1.166 M     & 2.307  \\
            \rowcolor[gray]{0.9} UltraFusion          & 1.860 B & 944.4  \\
            \rowcolor[gray]{0.9} Ours          & 1.872 B & 282.1 \\
            \bottomrule
            \end{tabular}%
        }
        \vspace{-8pt}
        \label{rebut_tab:flops}
    \end{minipage}
    \vspace{-1pt}
\end{table*}
\subsection{Quantitative results}
\label{subsec:quantitative}

\cref{tab:realhdrv_ultrabench} reports results on RealHDRV and UltraFusion Benchmark, which contain large exposure differences and diverse object motions. 
Our model consistently outperforms conventional approaches such as MEF-LUT~\cite{jiang2023meflut} and HSDS-MEF~\cite{wu2024hybridsupervised} across all quality metrics, while achieving comparable or even superior performance to the generative baseline UltraFusion~\cite{chen2025ultrafusion}. 
These results indicate that LIIFusion effectively handles occluded or saturated regions in dynamic scenes by adaptively integrating pixel-wise features through its implicit representation.
Moreover, our approach requires only one fourth of UltraFusion’s inference time, demonstrating that the proposed coarse-to-fine framework performs high-quality fusion far more efficiently than patch-wise generative models, even though lightweight methods such as MEF-LUT remain faster overall but fall short in quality.
As summarized in Table~\ref{rebut_tab:flops}, adding the fine fusion module slightly increases parameters over UltraFusion, but our coarse-to-fine design reduces TFLOPs by more than $3\times$ for a single image, preserving diffusion-driven quality while alleviating runtime bottlenecks.
\cref{tab:mefb} further presents results on the static MEFB dataset, where LIIFusion consistently outperforms baselines across all non-reference quality metrics. 
These results confirm that LIIFusion provides a strong balance between fidelity and efficiency, preserving generative-quality fusion while enabling faster inference for high resolution multi exposure imaging.

\begin{figure*}[t]
    \centering
    \begin{minipage}[t]{\linewidth}
        \centering
        \includegraphics[width=\linewidth]{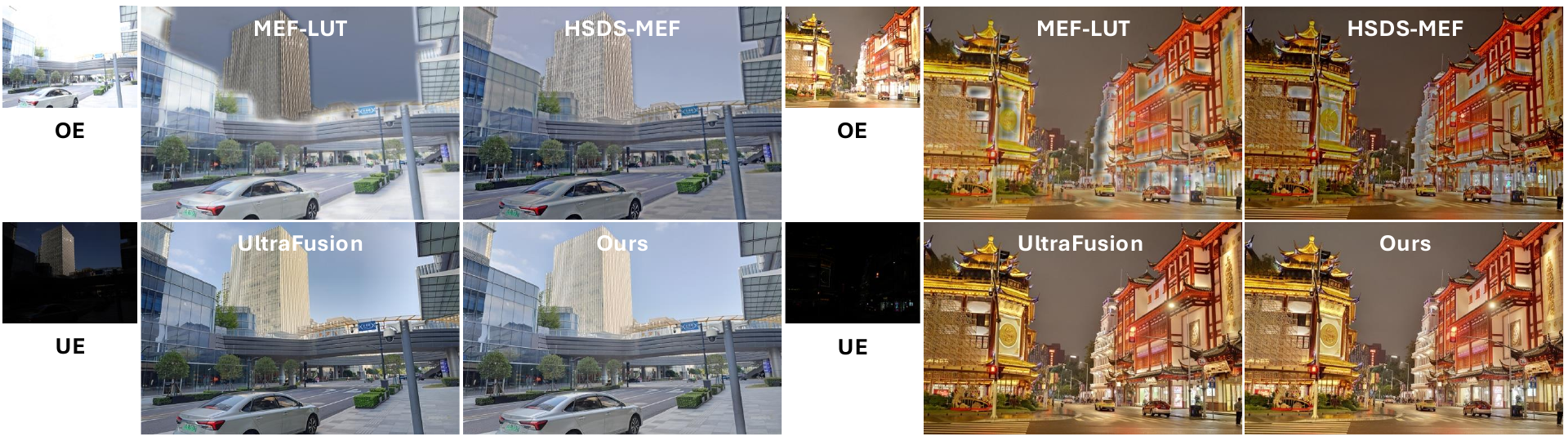}
        \captionof{figure}{
            Qualitative comparison on ultra HDR scenes in the UltraFusion Benchmark. Compared to prior MEF models, our approach achieves more stable exposure fusion, retaining structural detail and natural color appearance in both day and night scenes.
        }
        \label{fig:uhdr}
    \end{minipage}
    \hfill
    \begin{minipage}[t]{\linewidth}
        \centering
        \vspace{12pt}
        \includegraphics[width=\linewidth]{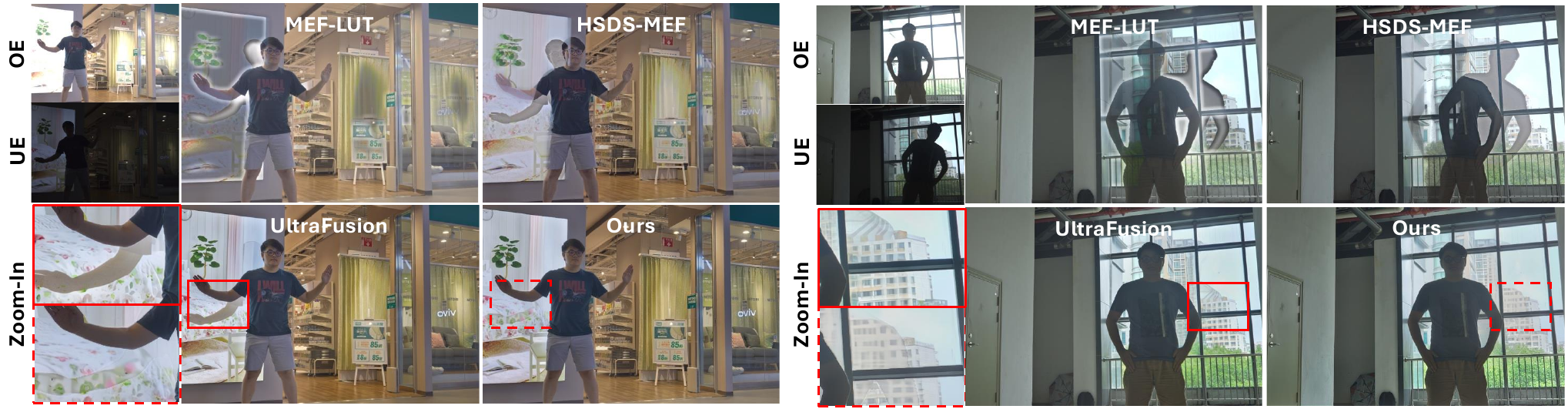}
        \captionof{figure}{
            Qualitative results on the RealHDRV dataset, showing that our method provides cleaner fusion with improved guided inpainting, robust occlusion handling, and reduced ghosting.
        }
        \label{fig:motion}
    \end{minipage}
    \vspace{-1pt}
\end{figure*}

\subsection{Qualitative results}
\label{subsec:qualitative}

We further conduct qualitative comparisons on ultra high dynamic range scenes from the UltraFusion Benchmark and dynamic motion scenes from the RealHDRV dataset. 
For fair comparison, we visualize results from MEF-LUT~\cite{jiang2023meflut}, HSDS-MEF~\cite{wu2024hybridsupervised}, UltraFusion~\cite{chen2025ultrafusion}, and our LIIFusion.
\newline
\\
\textbf{Ultra HDR scenes.}
As shown in \cref{fig:uhdr}, MEF-LUT produces inconsistent tone transitions due to its LUT-based fusion, resulting in sky–ground discontinuities and unnatural illumination in building regions. 
HSDS-MEF, though content-adaptive, fails to fully cover the full dynamic range and often produces dark or low-contrast results. 
Both UltraFusion and our method generate visually natural outputs while our approach achieves these comparable results at one fourth of UltraFusion’s inference time.
\newline
\\
\textbf{Dynamic motion scenes.}
\cref{fig:motion} presents results on large-motion scenarios. 
MEF-LUT and HSDS-MEF suffer from severe ghosting artifacts when aligning moving objects, while UltraFusion produces smoother results via guided inpainting but still struggles in extreme motion regions. 
Because UltraFusion operates on patch-wise latent inpainting, its contextual range is limited, sometimes leading to uncertain synthesis in occluded areas. 
In contrast, LIIFusion performs coarse fusion over the entire downsampled image, enabling globally consistent generation and effective utilization of diffusion priors. 
As shown in the highlighted regions, our model successfully removes ghost artifacts and restores missing structural details, confirming its efficiency and robustness in complex dynamic scenes.

\subsection{Ablation study}
All ablation studies except \cref{subsec:liif_ablation} are conducted on the UltraFusion Benchmark dataset to ensure a consistent evaluation environment across different configurations.

\subsubsection{Adaptive exposure correction ablation}
\label{subsec:aec_ablation}

We first evaluate the contribution of the proposed adaptive exposure correction (AEC).
As shown in \cref{tab:aec_ablation}, our method achieves higher overall image quality than UltraFusion regardless of whether AEC is applied. While the numerical gain from AEC is relatively modest, its qualitative impact is substantially more pronounced.

AEC is designed to address a core difficulty of low resolution generative fusion: saturated regions in over-exposed inputs often lack recoverable structure, causing the diffusion model to hallucinate blurred or distorted content during coarse fusion. To examine this effect, \cref{fig:aec_ablation} visualizes the low resolution MEF outputs before fine refinement. Without AEC, the model struggles to synthesize coherent structures in heavily saturated areas, where neon signs become blurred, map details on display boards collapse, and bright decorative lights lose their shape. These artifacts originate in the coarse stage and cannot be fully restored by the fine fusion module, which refines but does not reinterpret the underlying structure.

With AEC, however, the saturated regions are attenuated before entering the diffusion model, allowing the coarse generative stage to preserve the essential structure of challenging areas. As a result, the LR fusion already contains clearer letter shapes, map boundaries, and high-frequency lighting patterns. When passed to the fine-stage implicit refinement, these structurally reliable coarse predictions lead to high resolution fusion results where both the global scene layout and fine details remain well aligned and visually coherent.

Overall, AEC strengthens the structural reliability of the coarse generative fusion, enabling the fine stage to more effectively restore details and produce consistent HDR-like outputs.

\subsubsection{Comparison between LIIFusion and LIIF}
\label{subsec:liif_ablation}

\begin{figure*}[!t]
    \centering
    \begin{minipage}[t]{0.48\linewidth}
        \centering
        \includegraphics[width=\linewidth]{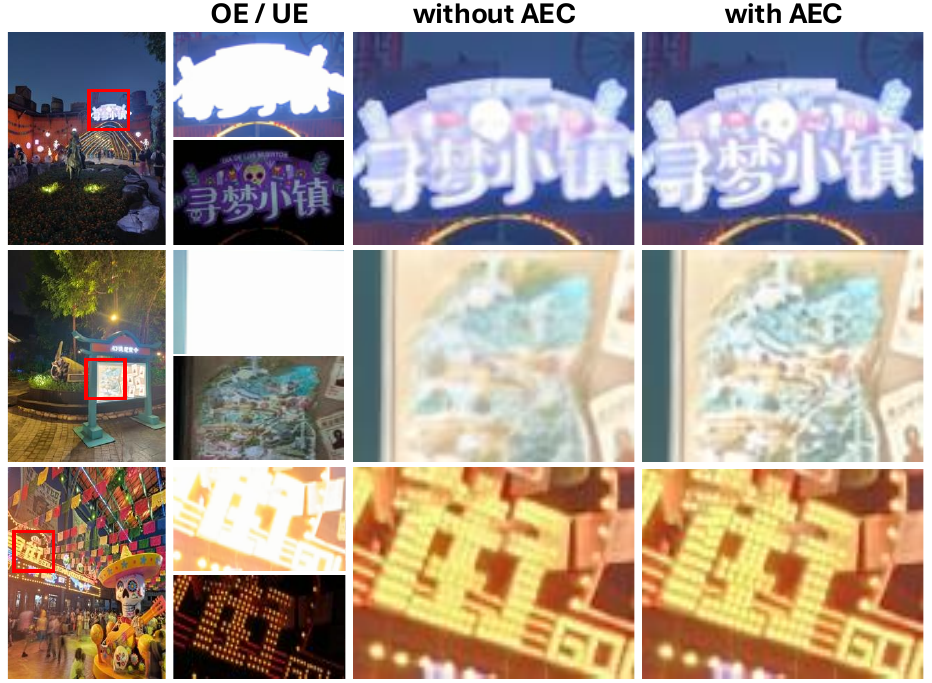}
        \captionof{figure}{
            Visual ablation of the adaptive exposure correction module on the UltraFusion Benchmark at low resolution, showing improved reconstruction of details lost in saturated OE regions.
        }
        \label{fig:aec_ablation}
    \end{minipage}
    \hfill
    \begin{minipage}[t]{0.48\linewidth}
        \centering
        \vspace{-135pt}
        \captionof{table}{Ablation study of the adaptive exposure correction.}
        \resizebox{\linewidth}{!}{%
        \begin{tabular}{l|cccc}
        \toprule
        \multicolumn{1}{c|}{Method}   & MUSIQ & PAQ2PIQ & HyperIQA & DeQA-Score \\ 
        \midrule\midrule
        UltraFusion & 68.40    & 75.18       & 0.6214 & \textbf{4.0123}       \\
        Ours (w/o AEC) & 70.07  & \textbf{75.64}   & 0.6451 & 3.9870          \\
        Ours (w/ AEC)           & \textbf{70.19}   & 75.59      & \textbf{0.6467} & 3.9807     \\ 
        \bottomrule
        \end{tabular}%
        }
        \vspace{-2pt}
        \label{tab:aec_ablation}

        \vspace{6pt}
        \captionof{table}{Ablation study of LIIF within the multi-exposure fusion framework. LIIF (pre-trained) is trained on DIV2K~\cite{agustsson2017ntire} and LIIF (from scratch) is trained on SICE dataset~\cite{cai2018learning}.}
        \resizebox{\linewidth}{!}{%
            \begin{tabular}{l|cc}
            \toprule
            \multicolumn{1}{c|}{Method}        & MEF-SSIM & HyperIQA \\ 
            \midrule \midrule
            LIIF (pre-trained)     & 0.8942     & 0.6389   \\
            LIIF (from scratch) & 0.9172    & 0.6230     \\
            Ours           & \textbf{0.9201}     & \textbf{0.6418}   \\ 
            \bottomrule
            \end{tabular}%
        }
        \vspace{-2pt}
        \label{tab:liif_ablation}
    \end{minipage}
    \vspace{-1pt}
\end{figure*}
\begin{figure*}[t]
    \centering
    \includegraphics[width=\linewidth]{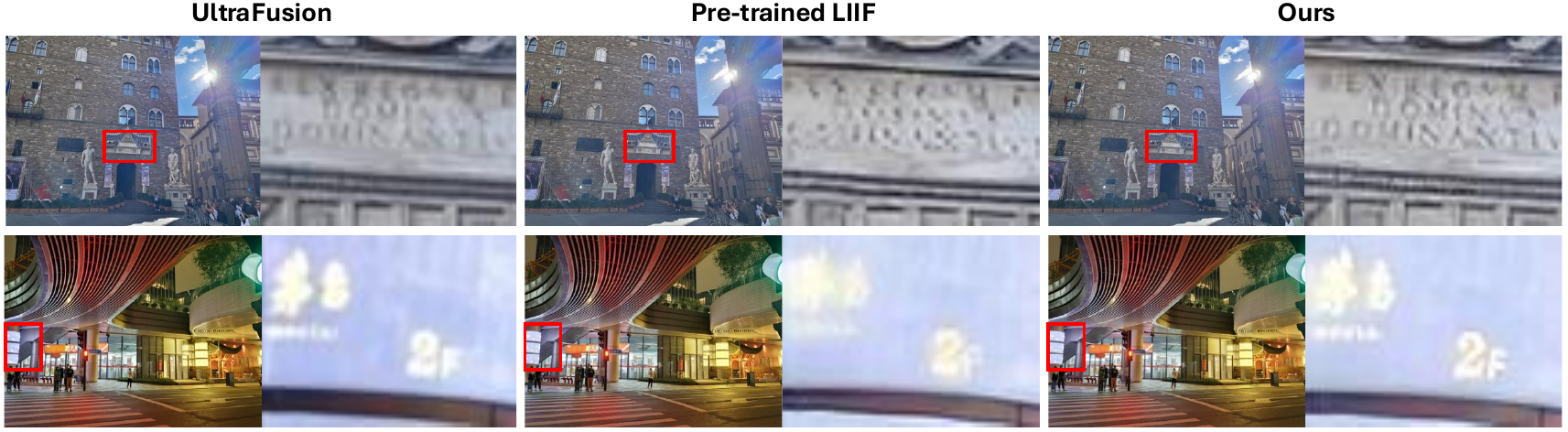}
    \caption{Visual comparison between the pre-trained LIIF and our LIIFusion on the UltraFusion Benchmark. Our LIIFusion restores finer textual and structural details, such as characters and patterns, which are blurred or lost in the pre-trained LIIF outputs.}
    \label{fig:liif_ablation}
    \vspace{-1pt}
\end{figure*}

\label{subsec:patch_size_ablation}
We further examine the effectiveness of adapting the LIIF architecture for MEF rather than conventional super resolution.
Tab.~\ref{tab:liif_ablation} compares three configurations: 
(1) the pre-trained LIIF trained on DIV2K~\cite{agustsson2017ntire}, 
(2) a LIIF model trained from scratch on SICE for super resolution, 
and (3) our proposed LIIFusion trained for multi exposure fusion. This ablation is performed on MEFB to measure structure consistency.

As shown in \cref{tab:liif_ablation}, the first two models achieve comparable results, indicating that conventional super resolution training, even with exposure data, does not provide sufficient cues for accurate fusion.
In contrast, our LIIFusion improves all metrics including MEF-SSIM for structural fidelity and HyperIQA for perceptual quality, demonstrating the benefit of learning to fuse multi exposure inputs rather than simply upsampling the coarse MEF result.

The qualitative comparison in \cref{fig:liif_ablation} further supports this observation. 
The pre trained LIIF models produce blurry textures and fail to recover fine details such as text and small structural elements.
UltraFusion, though structurally strong due to its generative prior, still lacks pixel-level consistency in high-frequency areas. 
Our model, on the other hand, synthesizes sharper and more coherent details, accurately reconstructing neon letters and map textures visible in zoomed-in regions. 
This confirms that framing LIIF as a fine stage fusion module, rather than as a simple super resolution network, enables precise pixel wise integration of exposure information, effectively complementing the coarse generative fusion stage.





\begin{figure*}[!t]
    \centering
    \begin{minipage}[t]{0.48\linewidth}
        \centering
        \includegraphics[width=\linewidth]{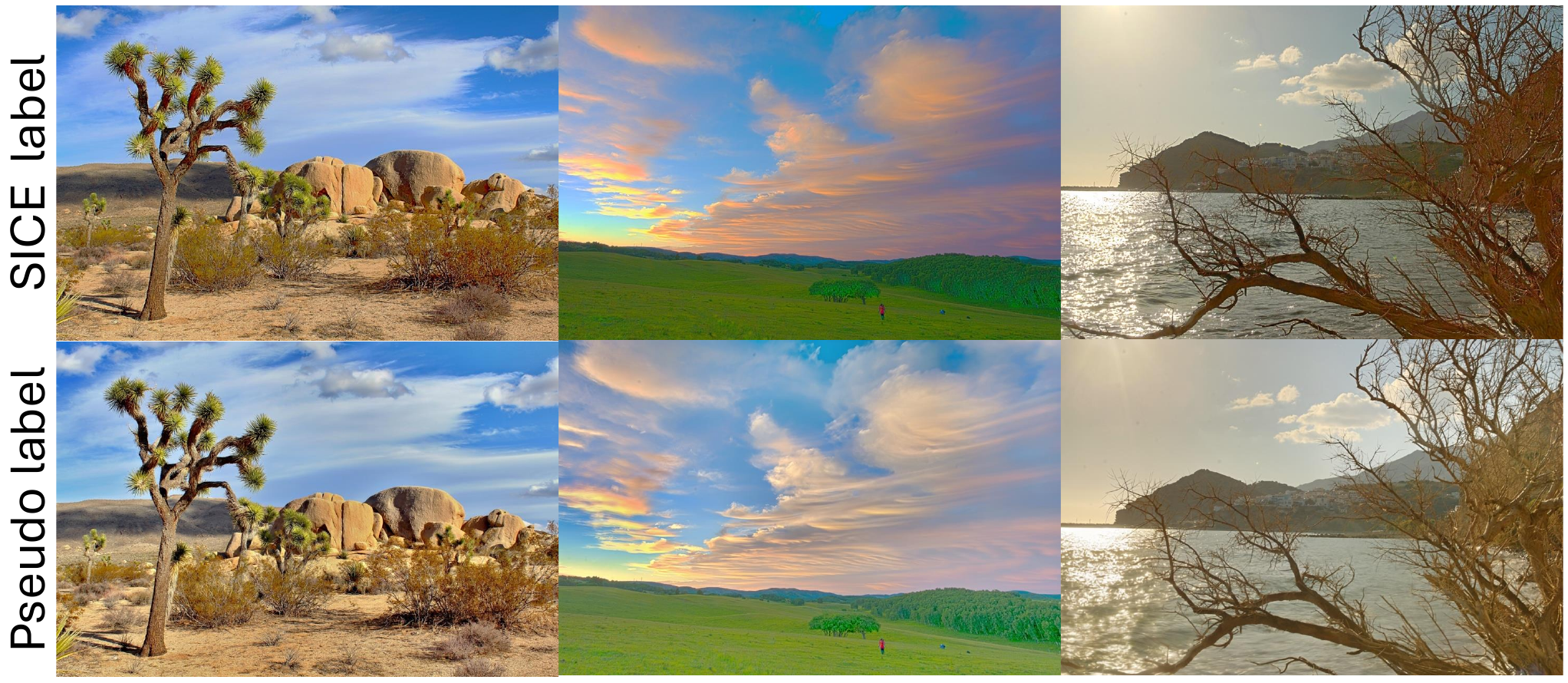}
        \captionof{figure}{
            Qualitative comparison between the original SICE labels and our generated pseudo labels.
        }
        \label{fig:dataset}

        \vspace{6pt}
        \captionof{table}{Quantitative comparison between using the original SICE label and our generated pseudo labels.}
        \resizebox{\linewidth}{!}{%
        \begin{tabular}{l|cccc} 
            \toprule
            \multicolumn{1}{c|}{Dataset} & MUSIQ & DeQA-Score & PAQ2PIQ & HyperIQA \\ 
            \midrule
            SICE label                  & 70.19     & 3.9807    & 75.59       & 0.6467        \\
            SICE pseudo label            & 69.34    & 3.9688    & 75.19    & 0.6323        \\
            \bottomrule
        \end{tabular}%
        }
        \vspace{-2pt}
        \label{tab:dataset_ablation}

        \vspace{6pt}
        \captionof{table}{Ablation study on exposure inputs for fine fusion. \textbf{Bold} denote the best performance and \underline{underlined} indicate the second best.}
        \resizebox{\linewidth}{!}{%
        \begin{tabular}{l|cccc}
            \toprule
             & MUSIQ & DeQA-Score & PAQ2PIQ & HyperIQA \\
            \midrule \midrule
            w/o OE, UE & 54.65 & 3.7961 & 74.09 & 0.4277 \\
            w/ UE only & 62.73 & 3.8412 & 74.59 & 0.5773 \\
            w/ OE only & \underline{68.38} & \underline{3.8438} & \underline{75.58} & \underline{0.6141} \\
            w/ OE, UE & \textbf{70.19} & \textbf{3.9807} & \textbf{75.59} & \textbf{0.6467} \\
            \bottomrule
        \end{tabular}
        }
        \label{tab:exposure_guidance}
    \end{minipage}
    \hfill
    \begin{minipage}[t]{0.48\linewidth}
        \centering
        \vspace{-75pt}
        \includegraphics[width=\linewidth]{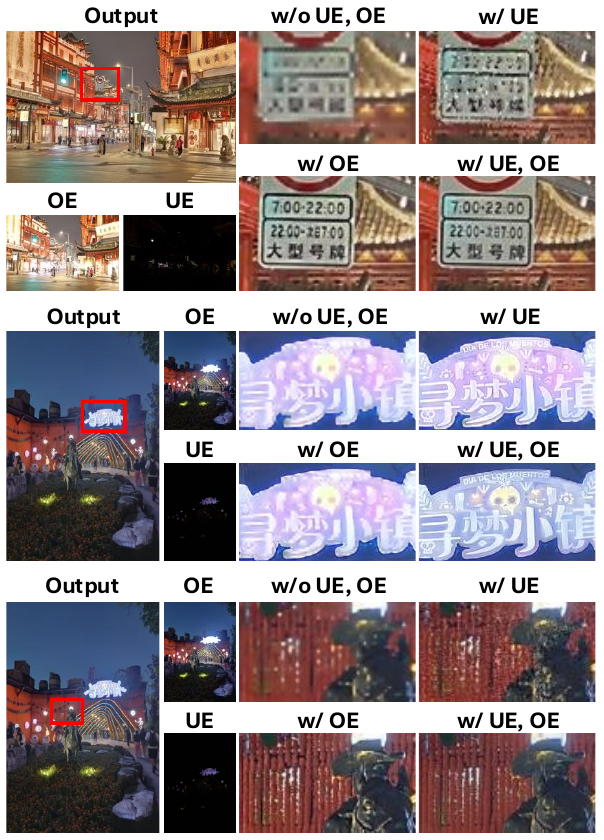}
        \captionof{figure}{
            Comparison of different exposure-conditioning strategies for LIIF. Using both OE and UE exposures preserves fine details and reduces residual artifacts compared with configurations using only one exposure or none.
        }
        \label{sup_fig:exposure_input}
    \end{minipage}
    \vspace{-2pt}
\end{figure*}

\subsubsection{Training dataset}
\label{subsec:dataset_ablation}

We additionally examine whether LIIFusion can be trained using pseudo labels instead of human-crafted ground truth. MEF datasets are difficult to scale because generating high-quality fused labels typically requires manual tone mapping or expert-designed fusion strategies. To explore whether synthetic annotations are sufficient for training, we replace the original SICE labels with the fused outputs generated by UltraFusion.

As shown in \cref{tab:dataset_ablation}, training with UltraFusion pseudo labels yields performance that is highly competitive with the model trained on the original SICE labels, with only marginal differences across all evaluation metrics. A qualitative comparison in \cref{fig:dataset} further illustrates that the pseudo labels maintain comparable visual fidelity to the provided SICE labels, capturing natural tones and detailed structures without noticeable degradation.

These results indicate that LIIFusion can be effectively trained using synthetic supervision alone, reducing the dependency on manually curated MEF annotations. This also suggests that stronger generative models can serve as scalable label generators, enabling future expansion of Multi Exposure Fusion datasets without additional human labeling effort.

\subsubsection{Ablation on exposure inputs for fine fusion}
\label{subsec:dataset_ablation}

To further examine whether LIIFusion performs true multi-exposure fusion rather than simple super-resolution, we analyze the contribution of each exposure input in the fine fusion stage. Specifically, we evaluate four input configurations for the LIIF module: (1) using neither OE nor UE, (2) using UE only, (3) using OE only, and (4) using both OE and UE. To isolate the effect of exposure cues, we replace the missing OE input with an all-white image and the missing UE input with an all-black image.

As summarized in \cref{tab:exposure_guidance}, performance consistently improves as more exposure inputs are provided. Using only the coarse LR MEF without any exposure cues results in a significant drop across all metrics, confirming that LIIFusion depends on exposure-conditioned refinement rather than purely upsampling the coarse output. Among the two exposures, OE contributes the most, likely because over-exposed images retain more structure in bright regions where LR MEF tends to blur details. UE also provides complementary information but has a smaller effect individually.

As shown in \cref{sup_fig:exposure_input}, when the fine fusion module receives only the coarse result, the input lacks sufficient detail and the output remains blurry. Using UE alone can help recover details in saturated regions (e.g., the second scene), but in the first and third scenes it tends to transfer UE noise. Using OE alone preserves HR details from the OE input, yet it cannot fully restore saturated regions. In contrast, using UE, OE, and the coarse result together yields the most coherent fusion. These results indicate that the fine fusion module adapts to the characteristics of each input and effectively integrates their complementary information to produce the final output.

\subsubsection{Additional analysis}

Additional experiments further support LIIFusion. \cref{tab:user_study} shows that users clearly prefer LIIFusion over conventional and generative MEF baselines, with the highest best-choice rate and the best average rank, while requiring much less time than UltraFusion. \cref{tab:downsample_ablation} shows that the coarse diffusion stage can operate at different input resolutions through the same sampling pipeline, with higher resolutions generally improving perceptual quality. However, the runtime increases substantially at larger coarse resolutions. We therefore use a fixed coarse resolution for practical efficiency and rely on the fine-stage implicit fusion function to reconstruct HR details from the original OE/UE inputs.

\begin{table*}[t]
    \centering
    \begin{minipage}[t]{0.49\textwidth}
        \centering
        \captionsetup{font=footnotesize}
        \captionof{table}{Human preference study. 22 participants ranked four anonymized methods. Inference time is measured on a NVIDIA RTX A5000.}
        \vspace{0.5mm}
        \resizebox{0.8\linewidth}{!}{
        \begin{tabular}{lccc}
        \toprule
        Model & Best-choice rate$\uparrow$ & Avg. rank$\downarrow$ & Time\\
        \midrule
        MEF-LUT    & 3.79\% & 3.57 & 0.5s \\
        HSDS-MEF   & 13.64\% & 2.89 & 2m 27s \\
        UltraFusion& 21.21\% & 2.02 & 12m 9s \\
        Ours       & \textbf{61.36\%} & \textbf{1.52} & 3m 23s \\
        \bottomrule
        \end{tabular}}
        \label{tab:user_study}
    \end{minipage}
    \hfill
    \begin{minipage}[t]{0.47\textwidth}
        \centering
        \captionsetup{font=footnotesize}
        \captionof{table}{Coarse fusion also supports resolution-agnostic inference. Results are reported on RealHDRV measured on a single H100.}
        \vspace{0.5mm}
        \resizebox{\linewidth}{!}{
        \begin{tabular}{lcccc}
        \toprule
        Resolution & MUSIQ$\uparrow$ & PAQ2PIQ$\uparrow$ & HyperIQA$\uparrow$ & Time \\
        \midrule
        256 $\times$ 256       & 65.03 & 72.91 & 0.5423 & 694 sec \\
        512 $\times$ 512       & 68.82 & 74.34 & 0.5963 & 630 sec \\
        768 $\times$ 768 (ours)      & 69.52 & 74.06 & 0.6175 & 949 sec \\
        1024 $\times$ 1024     & 69.58 & 74.17 & 0.6206 & 2018 sec \\
        \bottomrule
        \end{tabular}}
        \label{tab:downsample_ablation}
    \end{minipage}
    \vspace{-1pt}
\end{table*}

\section{Discussion and future work}

Although LIIFusion offers clear benefits in both efficiency and fusion quality, several limitations remain. As shown in \cref{fig:aec_ablation}, the final output is still influenced by the structural reliability of the coarse generative stage. In addition, despite the reduced computational cost, our approach is still slower than lightweight methods such as MEF-LUT, limiting its applicability in real-time scenarios. Finally, in scenes with extremely large occlusions or rapid motion, our model may still produce subtle ghost artifacts, suggesting the need for more motion-aware generative mechanisms.

Although LIIFusion is not designed for real-time MEF, its runtime can be viewed alongside recent diffusion-based low-level vision methods such as LEDiff~\cite{wang2025lediff}, UltraFusion~\cite{chen2025ultrafusion}, and SIED~\cite{jiang2025learning}, which address challenging scenarios where non-generative approaches remain limited. In this context, LIIFusion substantially lowers the practical cost of generative MEF. As shown in \cref{tab:realhdrv_ultrabench}, it is much faster than UltraFusion and achieves a runtime comparable to non-generative HSDS-MEF while preserving the advantages of generative MEF.


For future work, we will explore extending this coarse-to-fine formulation to broader low-level vision tasks beyond MEF. By using diffusion to establish reliable coarse structure and Local Implicit Image Functions to recover fine details from guided inputs, the framework offers a practical way to deploy generative priors at high resolution while reducing their computational burden.
\section{Conclusion}

In this work, we introduced LIIFusion, a novel coarse-to-fine framework for generative Multi-Exposure Fusion that combines diffusion-based global harmonization with pixel-level implicit refinement. Our approach addresses the fundamental trade-off between generative power and structural fidelity by performing coarse fusion at a low resolution using a diffusion prior, then restoring high-resolution details through a resolution-agnostic multi-exposure implicit fusion function conditioned on HR OE/UE sources.
This architecture effectively bridges the gap between global exposure alignment and fine-grained structural reconstruction. 
This is not a generic diffusion-plus-super-resolution cascade, but a task-specific decomposition of generative MEF into global latent-space fusion and local RGB-space HR evidence fusion. As the first INR-based MEF formulation, LIIFusion lets each query jointly use the coarse output and aligned OE/UE inputs, injecting original HR exposure evidence into the final image.
Extensive experimental results and ablation studies demonstrate that LIIFusion delivers fusion quality comparable to, and in many cases exceeding, state-of-the-art diffusion baseline while requiring only one-quarter of their inference time. By providing a more efficient yet powerful synthesis paradigm, we hope that this framework provides a practical step toward making generative Multi-Exposure Fusion more feasible for real-world, high-resolution imaging applications and sets a new precedent for the integration of INRs in generative workflows.
\section*{\ackname}
This research was supported by Artificial Intelligence Graduate School Program grant funded by Yonsei University (RS-2020-II201361), Samsung Research Funding \& Incubation Center of Samsung Electronics (SRFC-IT2501-02), Institute of Information \& communications Technology Planning \& Evaluation (IITP) grant funded by the Korea government(MSIT): Developing a Sustainable Collaborative Multi-modal Lifelong Learning Framework (No.RS-2022-II220113) and Artificial Intelligence Research Hub Project (No. RS-2024-00457882).

\clearpage

\renewcommand{\thesection}{\Alph{section}}
\renewcommand{\thefigure}{\Alph{figure}}
\renewcommand{\thetable}{\Alph{table}}
\setcounter{section}{0}
\setcounter{table}{0}
\setcounter{figure}{0}

\setcounter{page}{1}
\begin{center}
    {\large \textbf{Coarse-to-fine Framework for Generative MEF \\ via Implicit Neural Representation}}\\[1em]
    {\large Supplementary Material}
\end{center}

\section{Additional ablation study}
We provide additional ablation studies in this section. All experiments are conducted on the UltraFusion Benchmark~\cite{chen2025ultrafusion} except for \cref{sup_sub_sec:alpha_ablation}, and evaluation is performed using the non-reference image quality assessment metrics MUSIQ~\cite{ke2021musiq}, PAQ2PIQ~\cite{ying2020patches}, and HyperIQA~\cite{you2025teaching}. These studies complement the main paper by examining architectural design choices, encoder configurations, and the behavior of our multi-exposure fusion (MEF) modules under various settings.

\subsection{Synthetic data augmentation}
Building on our observation in Tab. 6 of the main paper that LIIFusion can be trained reliably using synthetic MEF results as pseudo labels, we further investigate whether this property can be exploited to expand the training data beyond SICE. Since the ICCV23 multi-exposure dataset~\cite{tel2023alignment} provides only exposure brackets without fused ground truth, we generate synthetic MEF outputs using UltraFusion and treat them as pseudo labels. We then retrain LIIFusion on the combination of the original SICE dataset and the ICCV23 pseudo-labeled set.

As shown in \cref{sup_tab:data_aug}, augmenting the training set with ICCV23 pseudo labels yields slight improvements in MUSIQ and HyperIQA, indicating that synthetic supervision remains effective even when the additional images are collected from a different dataset. This experiment suggests that LIIFusion can benefit from datasets without MEF ground truth by leveraging synthetic MEF generation, and highlights the potential of our framework as a practical tool for scaling training data in scenarios where ground truth fused images are not available.

\subsection{Backbone of coarse feature encoder}
We evaluate different backbone choices for the coarse encoder used to extract features from the low resolution fused image $I_\text{mef}^{LR}$. Following common practice in LIIF-based architectures, we test three representative encoders: EDSR-baseline~\cite{lim2017enhanced}, RDN~\cite{zhang2018residual}, and SwinIR~\cite{liang2021swinir}. \cref{sup_tab:backbone} compares their performance and inference time.

While all backbones exhibit comparable performance, SwinIR achieves the best overall trade-off. It achieves the highest scores on MUSIQ and PAQ2PIQ and remains competitive on HyperIQA, while also providing the fastest inference time. Based on this favorable balance of efficiency and quality, we adopt SwinIR as the coarse encoder in LIIFusion.

\begin{table*}[!t]
  \centering
  \begin{minipage}[t]{0.48\linewidth}
    \centering
    \captionof{table}{Effect of synthetic data augmentation.}
    \vspace{-0.9em}
    \resizebox{\linewidth}{!}{%
    \begin{tabular}{l|ccc} 
      \toprule
      \multicolumn{1}{c|}{Dataset} & MUSIQ$\uparrow$ & PAQ2PIQ$\uparrow$ & HyperIQA$\uparrow$ \\ 
      \midrule
      SICE label                  & 70.19 & \textbf{75.59} & 0.6467        \\
      + ICCV23 pseudo label       & \textbf{70.22} & 75.46 & \textbf{0.6506}        \\
      \bottomrule
    \end{tabular}%
    }
    \vspace{-1pt}
    \label{sup_tab:data_aug}

    \vspace{10pt}
    \captionof{table}{Ablation study on backbone of coarse feature encoder.}
    \vspace{-0.9em}
    \resizebox{\linewidth}{!}{%
    \begin{tabular}{l|cccc}
      \toprule
       \multicolumn{1}{c|}{Model} & MUSIQ$\uparrow$
       & PAQ2PIQ$\uparrow$ & HyperIQA$\uparrow$ & Time$\downarrow$ \\
      \midrule \midrule
      EDSR-baseline & 69.77 & \underline{75.52} & 0.6351 & 69 min \\
      RDN & \underline{70.08} & 75.51 & \textbf{0.6473} & 71 min \\
      SwinIR (Ours) & \textbf{70.19} & \textbf{75.59} 
      & \underline{0.6467} & 59 min \\
      \bottomrule
    \end{tabular}
    }
    \label{sup_tab:backbone}
  \end{minipage}
  \hfill
  \begin{minipage}[t]{0.48\linewidth}
    \centering
    \captionof{table}{Ablation study on $\alpha$ of adaptive exposure correction on the MEFB dataset. Across different $\alpha$ values, there is a trade-off between structural preservation and fusion quality, and we choose the optimal setting $\alpha=0.2$.}
    \vspace{-0.9em}
    \resizebox{\linewidth}{!}{%
    \begin{tabular}{c|cccc}
      \toprule
      Value of $\alpha$   & MUSIQ & PAQ2PIQ & HyperIQA & MEF-SSIM \\ 
      \midrule\midrule
      1.0          & 65.41 & 72.23 & 0.6017 & 0.8633       \\
      0.8           & 66.71  & 72.73 & 0.6100 & 0.9196          \\
      0.6           & 67.27  & 72.94  & 0.6181 & 0.9322     \\ 
      0.4           & 67.60  & 73.11  & 0.6227 & \textbf{0.9335}   \\ 
      \rowcolor[gray]{0.9} 0.2 (Ours)     & 67.91 & 73.28  & 0.6265 & 0.9313 \\ 
      0.0 (UltraFusion)  & \textbf{68.14} & \textbf{73.45} & \textbf{0.6309} & 0.9266       \\
      \bottomrule
    \end{tabular}%
    }
    \label{sup_tab:alpha_ablation}
  \end{minipage}
\end{table*}

\begin{figure}[t]
    \centering
    \includegraphics[width=\linewidth]{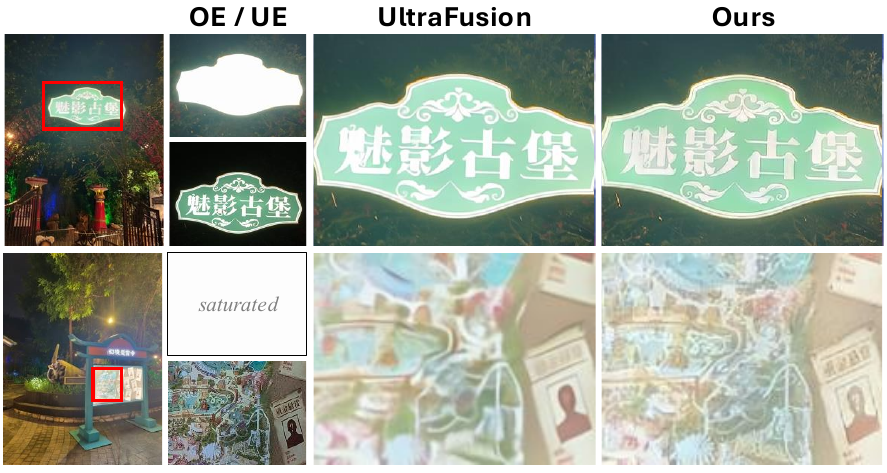}
    \vspace{-2em}
    \caption{
         Comparison of high resolution MEF results. With AEC and our fine feature encoder, LIIFusion preserves far more structure in saturated regions, recovering text and map details that become blurred or lost in UltraFusion.
    }
    \label{sup_fig:aec_ablation}
    \vspace{-1pt}
\end{figure}


\subsection{Ablation on adaptive exposure correction}
\label{sup_sub_sec:alpha_ablation}

The adaptive exposure correction (AEC) is designed to alleviate the structural degradation that occurs when performing multi-exposure fusion at low resolution. As shown in Fig. 5 of the main paper, saturated regions in the over-exposed (OE) image often lose fine structures, yielding blurry or collapsed details that the generative MEF model cannot reliably recover. By attenuating overly bright pixels prior to coarse fusion, AEC restores structure that would otherwise be lost in these regions.

To determine the optimal correction strength, we vary the hyperparameter $\alpha$, which controls Eq. (6) in the main paper. Larger values of $\alpha$ impose stronger suppression on saturated areas. We conduct this ablation on the MEFB dataset~\cite{Zhang2021} using the low-resolution MEF outputs, as these directly reflect the structural fidelity produced by the coarse stage.

\cref{sup_tab:alpha_ablation} shows that the non-reference image quality assessment metrics generally favor the setting without correction ($\alpha = 0.0$), consistent with the behavior of the original UltraFusion model. In contrast, the MEF-SSIM~\cite{ma2015perceptual} score, which captures structural consistency, improves whenever AEC is applied and reaches its best value around $\alpha = 0.4$. Considering both perceptual quality and structure preservation, we select $\alpha = 0.2$ as the default setting, which provides a balanced performance across all metrics.

As further illustrated in \cref{sup_fig:aec_ablation}, even mild correction enhances the reconstruction of saturated details such as neon signage and high-intensity textures. These results confirm that AEC strengthens the robustness of the coarse fusion stage and supplies more structurally reliable inputs for the subsequent fine fusion process.
\section{Implementation details for inference}

Our inference pipeline follows a coarse to fine procedure that integrates low resolution generative fusion with high-resolution implicit function based pixel level fusion. A central requirement of this pipeline is maintaining perfect pixel alignment between the low resolution fusion stage and the high-resolution fine fusion stage. The full inference process is described below.

\begin{figure}[t]
    \centering
    \includegraphics[width=\linewidth]{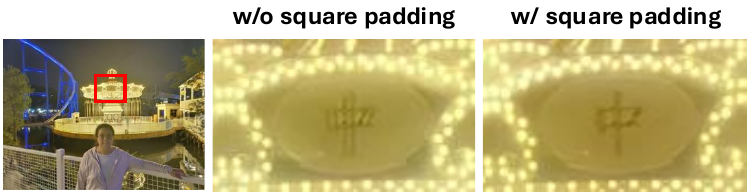}
    \vspace{-1em}
    \caption{
        Effect of square padding during inference. Without padding, subpixel misalignment leads to blurred fine details. With padding, alignment is preserved and details remain sharp.
    }
    \label{sup_fig:square_inference}
    \vspace{-1pt}
\end{figure}

We begin by applying zero padding to the high-resolution OE and UE images so that both become square. This step is critical for preserving consistent spatial scaling across axes. Without padding, resizing the shorter edge to a fixed size such as 512 leads to a non integer scale factor for the longer edge (for example, 773.3 becomes 773 after rounding). Even this fraction of a pixel introduces misalignment between the fused low resolution output and the original high-resolution exposures. Such sub pixel drift propagates into the fine fusion stage and results in ghosting artifacts as shown in \cref{sup_fig:square_inference}. Padding the images into a square ensures that both dimensions share the same integer scaling factor, preventing this issue.

The padded images are then downsampled to the target resolution (e.g., 512$\times$512 or 768$\times$768), producing low resolution OE and UE. Optical flow is computed between these two low resolution exposures, and the low resolution UE is backward warped to align with the low resolution OE. The same flow field is then upsampled and applied to the high-resolution UE image, ensuring alignment consistency between the low resolution and high-resolution branches.

Coarse fusion is performed using the aligned low resolution exposures through the generative MEF module. This stage resolves large exposure differences, suppresses occlusions, and generates globally coherent structures at a low computational cost due to the reduced spatial resolution.
The coarse fusion output, along with the high-resolution OE and the aligned high-resolution UE, is then processed by our LIIF based fine stage. This implicit function based pixel level fusion refines the coarse output by synthesizing high-frequency textures and restoring fine structural details that are not recoverable through generative fusion alone.

Finally, we remove the original zero padding and crop the fused output back to its original aspect ratio, producing the final high-resolution MEF result.
This inference strategy maintains strict pixel alignment across all stages, avoids sub pixel inconsistencies caused by uneven resizing, and effectively combines global generative reasoning with precise local pixel level fusion to produce high-quality multi exposure results.
\section{Justification of using LIIF in the fine stage}

We choose LIIF~\cite{chen2021liif} as the fine fusion module. This model choice is for efficient fusion guided by a long-standing super resolution (SR) trend : (i) avoid dense HR-grid computation and (ii) keep the refinement module resolution-agnostic. FSRCNN~\cite{dong2016accelerating} motivates ``process-then-upsample'' by showing that ``upsample-then-process'' becomes expensive on HR grids, reporting $>40\times$ speedup over SRCNN~\cite{dong2015image}'s upsample-then-process pipeline. LIIF takes the next step: instead of committing to discrete scales (e.g., $\times 2/\times 3/\times 4$, requiring scale-specific parameters), it represents the output as a continuous function queried at arbitrary coordinates, so one set of parameters supports any scale without retraining or storing multiple scale checkpoints. This directly matches MEF deployment, where target resolution varies by sensor and post-crop. Importantly, LIIF here is not a drop-in SR head: our ablations show that a pretrained (SR-trained) LIIF is insufficient for MEF, while LIIFusion training improves MEF metrics (MEF-SSIM / HyperIQA).
\section{Additional results}

\begin{figure}[t]
    \centering
    \includegraphics[width=\linewidth]{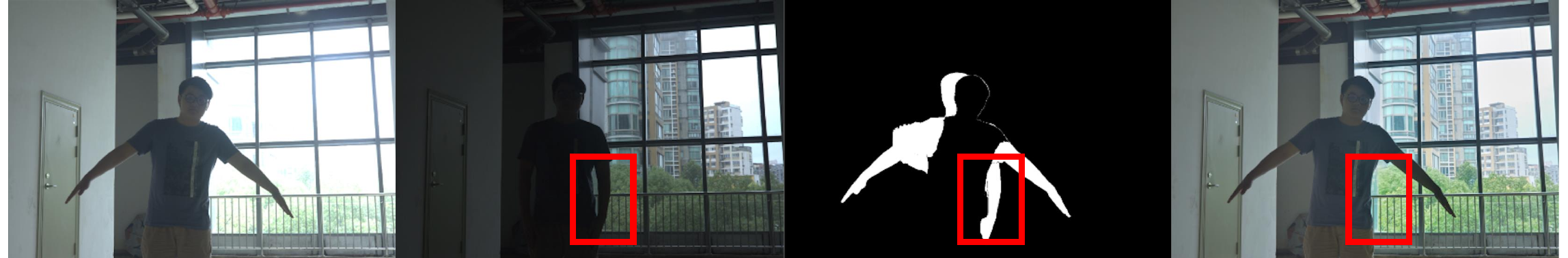}
    \captionsetup{font=footnotesize}
    \caption{
         Robust fusion with inaccurate flow-based masks. Our LR-domain flow estimation improves reliability (main paper Fig. 4), yet correspondences may still be imperfect under extreme exposure gaps. Even in such cases, our method maintains a stable fusion result.
    }
    \label{rebut_fig:failure_case}
\end{figure}

\begin{figure}[b]
    \centering
    \includegraphics[width=0.9\linewidth]{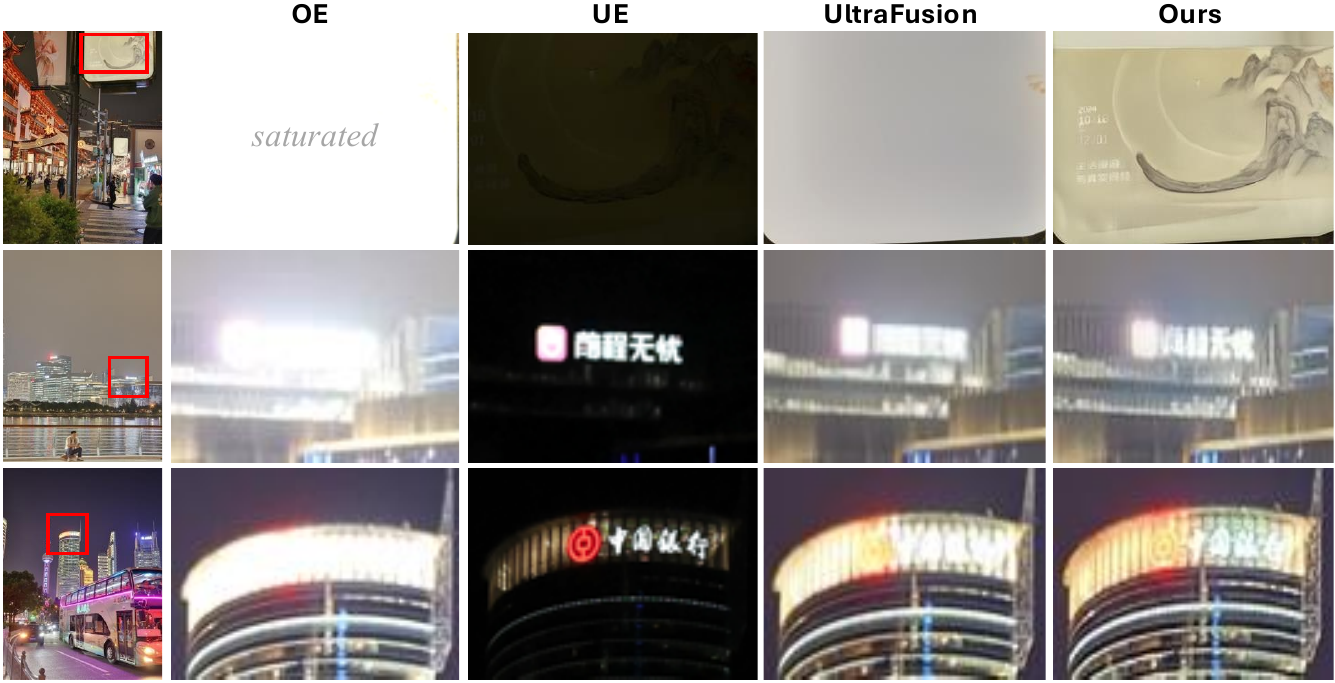}
    \caption{
        Comparison of additional qualitative results. The red-boxed regions highlight differences in detail preservation and appearance between UltraFusion and our method.
    }
    \label{sup_fig:ultra_vs_ours}
\end{figure}

\begin{figure*}[t]
    \centering
    \includegraphics[width=0.9\linewidth]{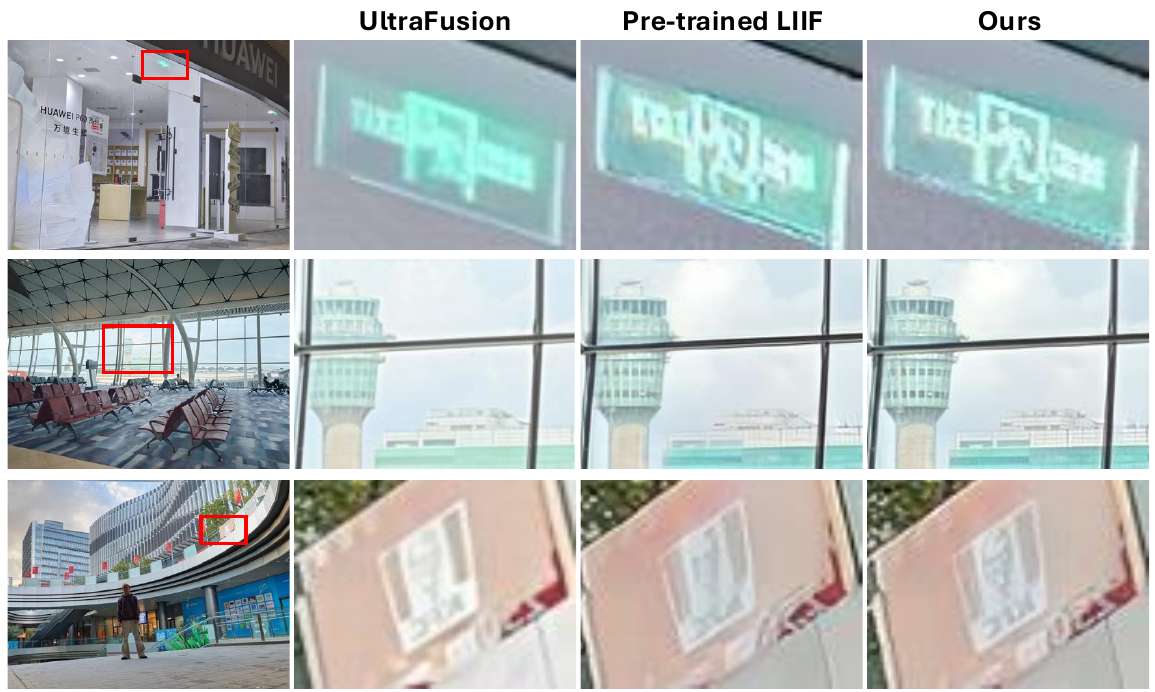}
    \caption{Comparison of additional qualitative results for LIIF. Our approach retains finer textures and more stable structural details across scenes.}
    \label{sup_fig:liif_ablation}
\end{figure*}

\begin{figure}[t]
    \centering
    \includegraphics[width=\linewidth]{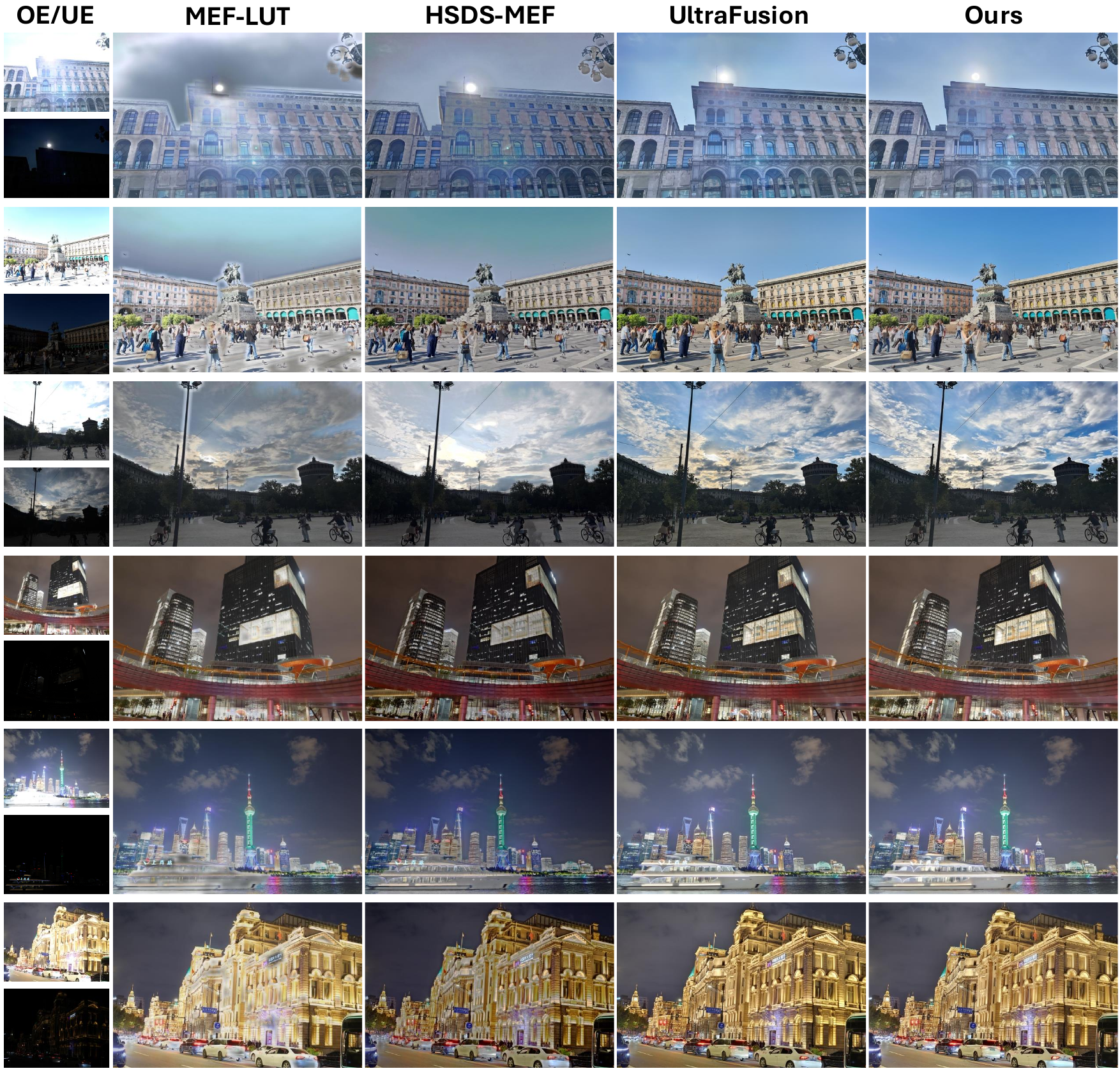}
    \caption{
        Comparison of additional ultra high dynamic range results, showing improved preservation of bright areas and enhanced contrast.
    }
    \label{sup_fig:uhdr}
\end{figure}

\begin{figure}[t]
    \centering
    \includegraphics[width=\linewidth]{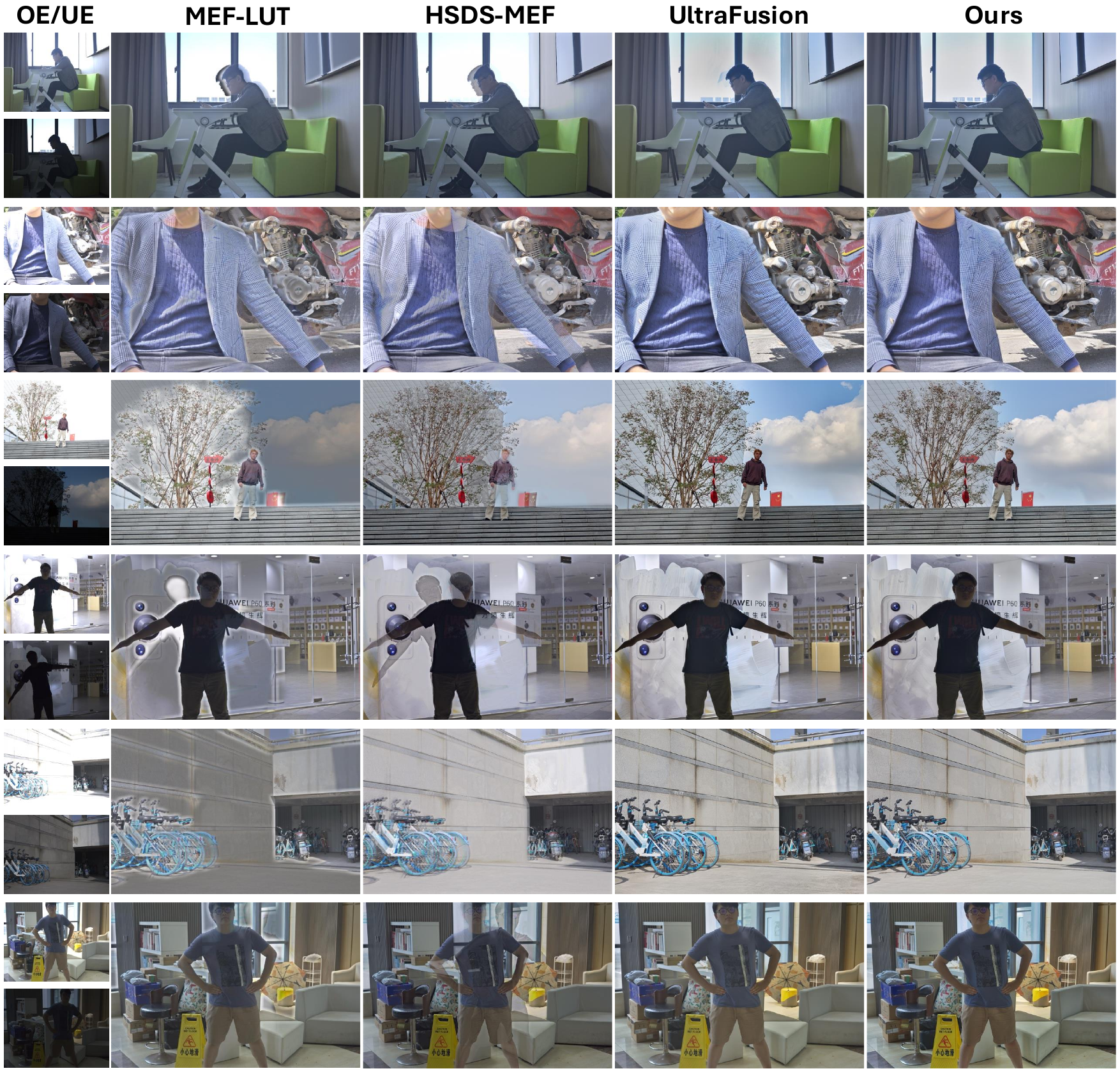}
    \caption{
        Comparison on additional multi exposure fusion results in dynamic scenes, showing that our method effectively suppresses motion-induced ghost artifacts while preserving scene details and contrast.
    }
    \label{sup_fig:dynamic}
\end{figure}

This section provides additional qualitative examples highlighting robustness, fine detail and ultra high dynamic range, presented in Figs.~\ref{rebut_fig:failure_case}-\ref{sup_fig:dynamic}.


\clearpage



%
%
\bibliographystyle{splncs04}
\bibliography{main}
\end{document}